\documentclass{article}
\usepackage{times}
\usepackage{graphicx} 
\usepackage{subfigure}

\usepackage{natbib}

\usepackage{algorithm}
\usepackage{algorithmic}

\usepackage{float}

\usepackage{hyperref}

\newtheorem{definition}{Definition}
\newtheorem{lemma}{Lemma}

\newtheorem{proposition}{Proposition}

\usepackage{amsmath}
\usepackage{amssymb}

\usepackage[accepted]{icml2019}

\icmltitlerunning{Towards Frequency-Based Explanation for Robust CNN}
\begin{document}

\twocolumn[
\icmltitle{Towards Frequency-Based Explanation for Robust CNN}



\icmlsetsymbol{equal}{*}

\begin{icmlauthorlist}
\icmlauthor{Zifan Wang*}{cmu}
\icmlauthor{Yilin Yang*}{cmu}
\icmlauthor{Ankit Shrivastava*}{cmu}
\icmlauthor{Varun Rawal*}{cmu}
\icmlauthor{Zihao Ding*}{cmu}
\end{icmlauthorlist}

\icmlaffiliation{cmu}{Carnegie Mellon University, Pittsburgh, PA 15213, USA}

\icmlcorrespondingauthor

\icmlkeywords{}

\vskip 0.3in
]



\printAffiliationsAndNotice{\icmlEqualContribution}  

\begin{abstract}
Current explanation techniques towards a transparent Convolutional Neural Network (CNN) mainly focuses on building connections between the human-understandable input features with models' prediction, overlooking an alternative representation of the input, the frequency components decomposition. In this work, we present an analysis of the connection between the distribution of frequency components in the input dataset and the reasoning process the model learns from the data. We further provide quantification analysis about the contribution of different frequency components toward the model's prediction. We show that the vulnerability of the model against tiny distortions is a result of the model is relying on the high-frequency features, the target features of the adversarial (black and white-box) attackers, to make the prediction.  We further show that if the model develops stronger association between the low-frequency component with true labels, the model is more robust, which is the explanation of why adversarially trained models are more robust against tiny distortions. 
\end{abstract}

\section{Introduction}
The gap between human's understanding and the logic behind Deep Neural Networks (DNNs) are receiving more attention as DNNs show competitive inference power in areas where humans used to be indispensable, e.g. medical diagnosis, auto-piloting, and credit systems. Entrusting users by explaining models' behavior becomes as necessary as promoting the performance. Explanations of deep neural networks tend to provide human-understandable descriptions about models' behaviors. Recent work of explanations focuses on associating importance of input features either to the model's output (e.g. \citet{karpathy2015visualizing}, \citet{sundararajan2017axiomatic}, \citet{Selvaraju_2019}) or to the distribution of data with which the model is trained (e.g. \citet{koh2017understanding}, \citet{yeh2018representer}, \citet{leino2018influencedirected});  Among all behaviors of deep models, the robustness draws increasing attentions due to the security and fairness concerns. Current discussion of models' robustness focuses on the capacity of defending adversarial attacks which aims to fool the deep models with similar input but the change is in-perceptional to humans\cite{szegedy2013intriguing}. Different adversarial attacks, though varying in the adversarial loss and updating rules, have been proved to successfully fool the most of deep models. (\citet{szegedy2013intriguing}, \citet{goodfellow2014explaining},  \citet{kurakin2016adversarial}, \citet{7958570}, \citet{7467366}).

Besides the work that aims to generate more robust model against existing attacks (\citet{aleks2017deep}, \citet{shafahi2019adversarial}) or a certifiable robust mechanism(\citet{cohen2019certified}), the community starts to explain why the deep model tends to be more vulnerable and tries to understand what is behind the adversarial attacks.\cite{aleks2017deep} concludes that a robust model tends to be more complex than the standard trained model and \citet{ilyas2019adversarial} shows that adversarial examples can be treated as an input full of non-robust features, which are poorly correlated to the labels. Models can still have good performance and generalizations by only learning the non-robust features but will become vulnerable against various attacks.

However, the community may overlook another facet of input data -- the distribution of frequency components. Treating inputs as signals in $\mathbb{R}^n$ space, we can decompose them into basis signals of diverse frequencies. While humans may only respond to a specific range of frequencies, DNNs are capable of using information from all frequencies. In this paper, we build the bridge between the model's robustness with the distribution of frequency components in the input set. Our major contributions are:
\begin{itemize}
    \item{} We prove that when the distortion is measured with $\ell_1$ distance, perturbing high-frequency components causes smaller change compared with the low-frequency components.
    \item{} We show that, for many existing adversarial attacks, there are more distortions in the high-frequency components. 
    \item{} We propose, Occluded Frequency, our measurement of the contribution of each frequency component towards the prediction and we further show that robust models are usually less relied on the high-frequency components in the input; therefore, they are more robust.
\end{itemize}
\section{Background and Related Work}
\label{sec: background}


Besides in the original feature space, several works provide new insights into the CNN behaviors from the aspect in the frequency domain. \citet{Haohan-2020} shows that unlike human beings, high-frequency components play significant roles in promoting CNN's accuracy. Adopting information from high-frequency components may cause the model to form very different concepts in learning as humans do. By observing adversarial defended models, \citet{Haohan-2020} concludes that smoothing the CNN kernels helps to enforce the model to use features of low frequencies. While the conclusion remains questionable due to the lack of theoretical proof is discussed, the paper proposes a novel view of attributing the frequencies components to the model's predictions. An alternative view of understanding the importance of frequency components is to observe a model's behavior when specific frequency components are modified. This area is studied known as the frequency components analysis on adversarial examples. \citet{guo2018low} proposes an adversarial attack only targeting the low-frequency components in an image, which shows that the model does utilize the features in the low-frequency domains for predictions instead of only learning from high-frequency components. \citet{Sharma-2019} demonstrated that state-of-the-art defenses are nearly as vulnerable as undefended models under low-frequency perturbations, which implies current defense techniques are only valid against adversarial attack in the high-frequency domain. On the other side, \cite{rahaman2018spectral} shows that ReLU networks tend to learn low-frequency featuers first and then pick up the high-frequency components later.

\subsection{Notation and Preliminary}
\label{sec: pre}
We first introduce the notation we are going to use in the rest of the paper. 

\noindent \textbf{Notation} A deep neural network $y = \arg\max_c f_c(\mathbf{x})$ that takes an input $\mathbf{x} \in \mathbb{R}^d$ and outputs a prediction class $y$. For simplicity, we omit the bold font and denote the input as $x$. We denote the $\ell_p$ norm as $||\cdot||_p$. The Discrete Fourier Transform of $x$ is denoted as $X = \mathcal{F}(x)$ (discussion to follow).

\noindent \textbf{Discrete Fourier Transform (DFT)} By convolution with a series of complex-valued exponential functions, DFT transforms a finite signal into the a complex-valued function of frequency. DFT is widely used in signal processing to analyze the frequency components for signals and human are more perceptional to the low frequency components in the signal, e.g. the content of image, while less perceptional to the high-frequency patterns, e.g. noise and small perturbations. Computing DFT requires dealing with complex function and we avoid this by introducing Discrete Cosine Transform in the real implementation but the analysis is still under DFT~\cite{bracewell1986fourier}.

\noindent \textbf{Discrete Cosine Transform (DCT)} DCT is similar to the discrete Fourier transform: it transforms a signal or image from the spatial domain to the frequency domain, but only maintaining the real part. The difference is the basis function: DFT uses complex exponential functions while the DCT uses real-valued cosine functions~\cite{1094144}. In the analysis part, we use DFT to demonstrate the motivations and methods while for the experiments we will replace DFT with DCT due to the imaginary components of DCT will bring extra computational complexity.

\subsection{Adversarial Attacks}

An adversarial attack tries to find a neighbor of an input x whose prediction is different from x but the change is in-perceptional to humans, causing failure in the reasoning. We introduce a few representative adversarial attack algorithms and methods. We discuss the three white-box attacks to be evaluated in this paper.  \citet{goodfellow2014explaining} proposes  \textbf{FGSM attack}, a heuristic searching for the adversarial examples following the sign of adversarial directions with a baby step at each time. It is usually done in the $\ell_\infty$ space by clipping the value outside the user-defined pixel range. Unlike FGSM, \textbf{PGD}~\cite{aleks2017deep} projects the adversarial samples learned from each iteration into the $\ell_p$ ball of the original input, therefore, the adversarial perturbation size is than the maximum allowed perturbation. \textbf{CW attack} \citet{7958570} reformalizes the adversarial loss to ensure that the solution to the optimization is close to the global optimal. They also solve the perturbation in the $\tanh$ space to yield the smoothness of the gradient signal. Besides the white-box attacks, we are also interested in the black-box attack and we include \textbf{SimBA} \cite{simba} in this paper. Under untargeted attack mode, SimBA tries to add a random perturbation on the input image at each step, and accept the perturbation if it decreases prediction certainty on the correct label.



\section{Adversarial attack analysis in frequency domain}

\label{sec: adv_attck}
The discussion of robustness is caused by the introduction of adversarial attacks and then the vulnerability comes into the attention. In the time domain, e.g. the image, adversarial perturbations are usually hard to discovered by humans since they are designed to have very small $\ell_p$ norms. It is natural to believe the adversarial perturbations do not distort the low-frequency components in the image but more like a change in the high-frequency pattern. We verify this hypothesis by introducing the following proposition.
\begin{proposition}[Lower Bound of Input Perturbation]
\label{prop: LBIP}
 Given an input $x$ and a perturbed input $x'$, the distortion measured in the $\ell_1$ space is lower bounded by the corresponding lowest frequency components. Formally,
 \begin{equation}
     ||x-x'||_1 \geq |X_0 - X'_0 |
 \end{equation}
\end{proposition}
\textit{Proof}:
We first write down the DFT $X = \mathcal{F}(x)$ of an given input $x$ and the inverse DFT $x = \mathcal{F}^{-1}(X)$.
\begin{equation}
\begin{aligned}
    X_k &= \mathcal{F}(x)_k = \sum^{d-1}_{i=0} x_i e^{-j\frac{2\pi}{d}ki}\\
x_i 
&=  \mathcal{F}^{-1}(X)_i = \frac{1}{d}\sum^{d-1}_{k=0} X_k e^{j\frac{2\pi}{d}ki} 
\end{aligned}
\end{equation}{}
where $j = \sqrt{-1}$ and $d$ is the dimension of $x$. And we also introduce Lemma \ref{lemma: exp sum}.
\begin{lemma} 
\label{lemma: exp sum}
The finite sum of complex-valued exponential series can be written as
\begin{equation}
\begin{aligned}
\sum^{d-1}_{i=0} e^{jix} &=
    \frac{1-e^{jdx}}{1-e^{jx}}
= \frac{-e^{idx/2(e^{-jdx/2}-e^{jdx/2})}}{-e^{jx/2}(e^{-jx/2}-e^{jx/2})} \\
&=\frac{\sin(dx/2)}{\sin(x/2)}e^{jx(d-1)/2}
\end{aligned} 
\end{equation}Let $x=\frac{2\pi}{N}k$ then we have the Fourier basis, so 
\begin{equation}
    \begin{aligned}
\sum^{d-1}_{n=0} e^{j\frac{2\pi}{d}ki} = \frac{\sin(\pi k)}{\sin(\frac{\pi}{d}k)}e^{jk\pi\frac{d-1}{d}} 
    \end{aligned}
\end{equation} Observe that 
$\sum^{d-1}_{n=0} e^{j\frac{2\pi}{d}ki}=d$ if $k=0$ and $0$ otherwise.
\end{lemma}

Eventually, we prove the Prop \ref{prop: LBIP}
\begin{equation}
\begin{aligned}
||x-x'||_1 &= \sum^{d-1}_{i=0} |x_i - x'_i|\\&=\sum^{d-1}_{i=0} |\mathcal{F}^{-1}(X)_i - \mathcal{F}^{-1}(X')_i|\\
&=\sum^{d-1}_{i=0} |\frac{1}{d}\sum^{d-1}_{k=0} X_k e^{j\frac{2\pi}{d}ki} - \frac{1}{d}\sum^{d-1}_{k=0} X'_k e^{j\frac{2\pi}{d}ki}|\\
&=\frac{1}{d}\sum^{d-1}_{i=0} |\sum^{d-1}_{k=0} (X_k- X'_k) e^{j\frac{2\pi}{d}ki}|\\
&\geq \frac{1}{d}|\sum^{d-1}_{i=0} \sum^{d-1}_{k=0} (X_k- X'_k) e^{j\frac{2\pi}{d}ki}|\\
&= \frac{1}{d}| \sum^{d-1}_{k=0} (X_k- X'_k) \sum^{d-1}_{i=0} e^{j\frac{2\pi}{d}ki}|\\
&=|X_0 - X'_0| \quad \text{(Use Lemma \ref{lemma: exp sum})}
\end{aligned}
\end{equation}

\noindent \textbf{Observation of Prop. \ref{prop: LBIP}} Perturbation applied to the input can be viewed either to a subset of features in the time domain or in the frequency domain by transforming the perturbation with DFT or DCT. For the same amount of perturbation measured by spectral energy in the frequency domain, Prop. \ref{prop: LBIP} shows that only the perturbation towards the low-frequency components will increase the lower bound of the same perturbation applied in the time domain. Therefore, attacking the low-frequency components of the input tends to cause higher distortion in the time domain due to the increase of lower bound, while the high-frequency distortion does not change the lower bound. As long as we require the distortion as small as possible in the time domain, an attacker should aim for mess up the high-frequency components instead of the low-frequency part, which matches our intuitions that adversarial perturbations tends to be not perceptional to human.

As the perturbation in the input space tends to happen to the high-frequency components, if we minimize the correlation between the model's decision and the use of high-frequency features, we should be able to force the attacker to make greater distortions than before. In the other word, the model will be more robust. To validate the proposition from empirical results, we propose a new way to measure the contribution of each frequency components towards the prediction.

\begin{figure*}[!t]
    \centering
    \includegraphics[width=\textwidth]{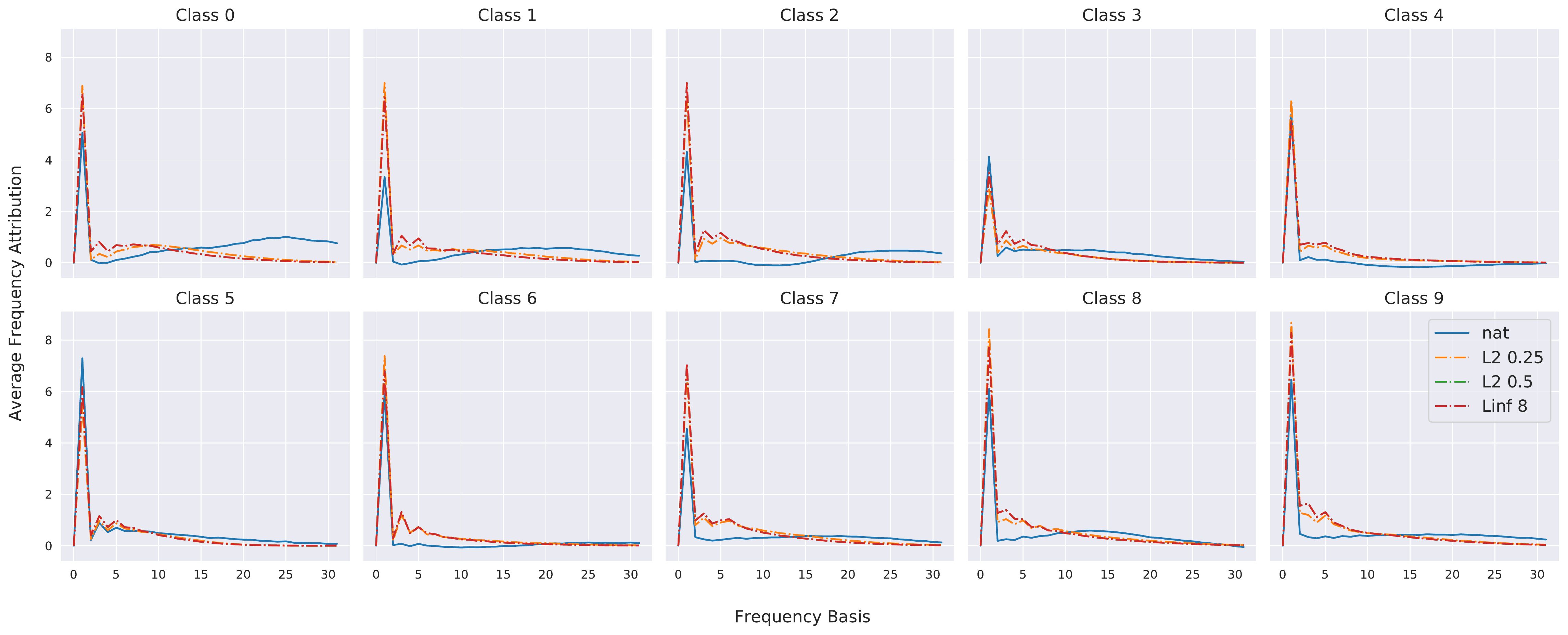}
    \caption{Average Attribution scores for each frequency components on each subset of CIFAR-10. We compute the attribution scores on three ResNet models: natural training (blue), Madry's training with $\epsilon=0.25$ in $\ell_2$ space (orange) and Madry's training with $\epsilon=0.5$ in $\ell_2$ space (green). Robust models tend to shift the high attribution scores from the high frequency range to the low frequency range, compared to the naturally trained models. Better viewed in color.}
    \label{rig: freq_attr}
\end{figure*}

\subsection{Contribution of Frequency Components}
Towards the interpretation of feature importance, one of the methods is attribution functions. Attribution functions aim to assign a score for each feature in the input and higher scores reflect higher relevance of the particular features towards the quantity of interest, e.g. the prediciton result of the model. One of the attribution methods is Occlusion \cite{zeiler2013visualizing} that ablates a subset of features in the input and assigns the importance score with the change of output compared to the original input. Ablation is usually performed by replacing the features with random noise or a baseline feature, e.g. zeros. We adapt the concept of Occlusion into the frequency domain so that the attribution score of each frequency component is computed by the change of the output of the model. Formally, we propose Occluded Frequency as an attribution method for the spectrum of input. 

\begin{definition}[Occluded Frequency (OF)]
Given a model $y = f(x)$, a class of interest $c$, define $H(i)$ as a matrix with the same shape of $\mathcal{F}(x)$ whose each entry $h_k$ is
\begin{equation}
    h_k = \mathbb{I}[k=i]\sigma(\mathcal{F}(x)_k) + \mathbb{I}[k\neq i]\mathcal{F}(x)_k
\end{equation}where $\sigma(\cdot)$ is a function that transforms an frequency component $\mathcal{F}(x)_k$ to a baseline version, e.g. zeros or random signal. Therefore, the occlusion score $ O(\mathcal{F}(x)_i)$ for frequency component $\mathcal{F}(x)_i$ towards class $c$ is defined as
\begin{equation}
    O(\mathcal{F}(x)_i) = f^c(x) - f^c(\mathcal{F}^{-1}(H(i)))
\end{equation} where $f^c(\cdot)$ is the logit score of class c.
\end{definition} 

\noindent \textbf{Ablation with DCT} To perform OF attribution on images with DCT, the masking is performed from the top left corner to the right bottom following the direction of increasing frequencies. All components with the same frequencies are ablated at the same time.

\noindent \textbf{Why Not Use Gradient} Another intuitive choise is to use the gradient-based attribution methods~\cite{simonyan2013deep, sundararajan2017axiomatic, shrikumar2017learning} to find the feature importance. However, gradient-based methods are shown to be non-robust and can cause contradictory results for tiny perturbations~\cite{ghorbani2017interpretation}. We find the ablation-based attributions are more consistent empirically, therefore, the result is more trustworth.

\section{Experiment}
\label{sec: exp}
In this section, we will use the retrained ResNet-50~\cite{he2015deep} to perform classification on CIFAR-10. For the robust model, we use pretrained model with adversarial examples\footnote{Weights: https://github.com/MadryLab/robustness}  \cite{robustness}. 

\subsection*{Experiment I: natural $\rightarrow$ adversarial}

\begin{figure}[h]
    \centering
     \includegraphics[width=0.5\textwidth]{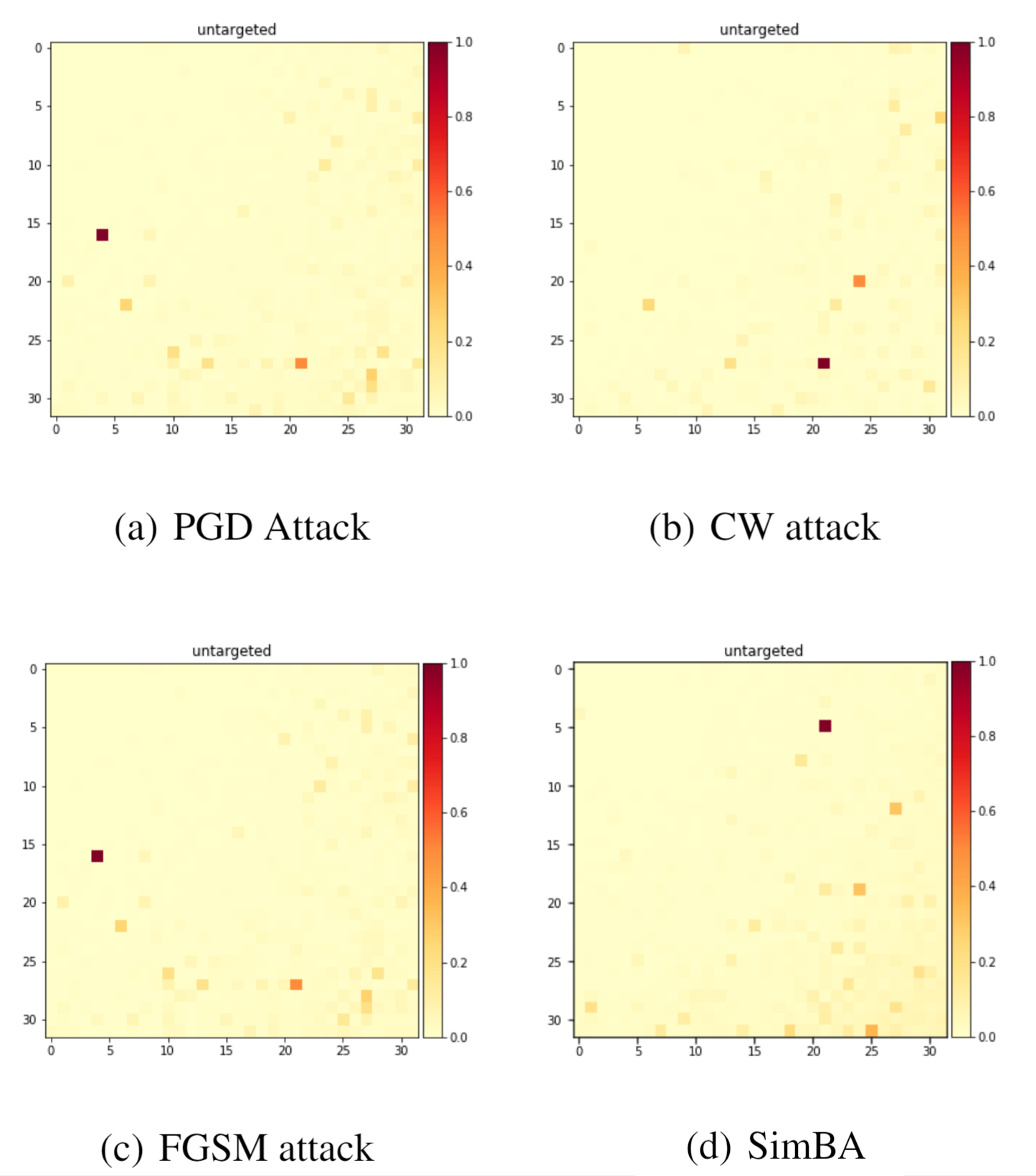} 
    \caption{RCT maps for different adversarial attacks on 200 images from CIFAR-10. The upper left corner and the lower right corner represent the lowest and highest frequency components in the DCT space, respectively. The deeper color indicates a greater change for a specific frequency component between the original images and the adversarial examples. RCT maps show that greater changes happen to the high frequency ranges. }
    \label{fig:ARCT}
\end{figure}

We first show the frequency domain of the adversarial examples generated by all methods mentioned in Sec \ref{sec: pre}. For PGD and FGSM, we clip the distortion in the $\ell_\infty$ ball with the maximum allowed perturbation $0.15$. In CW attack, we project the distortion in the $\ell_2$ ball. For SimBA, we used step size of 0.2 and we clip the distortion in the $l_{\infty}$ ball with the maximum allowed perturbation $0.2$. All the attacks are untargetted. The effect of adversarial attacks in the frequency domain for input is analyzed by calculating average Relative Change in discrete cosine Transforms (RCT) of an input $x$ and its perturbed image $x'$.
\begin{equation}
    \text{RCT} = 
\frac{1}{\text{N}}\sum_{\text{i}=1}^{\text{N}}|\frac{\text{DCT}(x_\text{i}') - \text{DCT}(x_\text{i})}{\text{DCT}(x_\text{i})}|
\end{equation} where $\text{DCT}(x)$ is discrete cosine transform of input $x$ and N is the number of samples in the dataset. We evaluate 200 images from CIFAR-10 dataset on a pre-trained VGG-19 model and the result is show in Figure \ref{fig:ARCT}. 

Fig \ref{fig:ARCT} show that all adversarial attacks, regardless of the white-box and the black-box, perturb images mostly at middle and high frequencies and low frequency perturbations are very small. This explains why perturbations are not detected by human eye whereas most of the deep network models give wrong predictions. 


\subsection*{Experiment II: natural $\rightarrow$ robust}
\begin{figure}[t]
    \centering
    \includegraphics[width=0.5\textwidth]{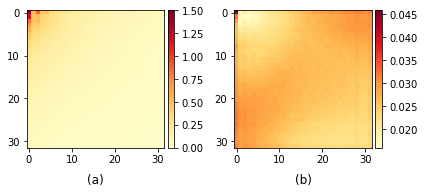}
    \caption{Frequency domain of the different between robust/non-robust images and the starting images. The left one (a) is the robust images and the right one (b) is the non-robust images}
    \label{fig:robust-dif}
\end{figure}

In this experiment, instead of finding the difference between a natural image with its adversarial counterpart, we explore the difference between with its robust counterpart. We use the definition of \emph{robust dataset} introduced by \cite{ilyas2019adversarial}, which attempts to find a neighbor $x_r$ of the original input $x$ where the features in $x_r$ remains correlated to the output within a certain maximum allowed perturbation. We solve the following objective to find the robust counterpart $x_r$. 
\begin{equation}
    x_r = \arg\min_{x_r} || g(x_r) - g(x) ||_2 \label{robust-opt}
\end{equation}

where x is the original image and $g(\cdot)$ is the mapping to a representation layer. When $g(\cdot)$ is extraced from a adversarially trained network, the generated dataset is robust. When the representation mapping $g(\cdot)$ is extracted from the standard network, the generated dataset is non-robust. The initial value for $x_r$ is an randomly selected image from the dataset. We conducted this generation process on the CIFAR-10 test dataset. We then show the difference in the Fourier domain between the robust counterpart and the non-robust counterpart with each natural image, respectively and the result is shown in Fig.\ref{fig:robust-dif}. The result demonstrates that the robust counterpart is mostly different from the original input in the low-frequency components while the non-robust counterpart is different on the middle-frequency components. Since the robust counterparts are generated from a robust model, we can also draw conclusions that a robust model only associates the low-frequency features with the labels while a regular model does not have a strong connection between the low-frequency features and the label. Instead, it captures a lot more features in the middle-frequency range.



\subsection*{Experiment III: Visualization of Contribution}
We have shown comparison analysis in Experiment I \& II on the different behaviors in the frequency domain for high-frequency features and low-frequency features. In this experiment, we show numerical analysis to further differentiate the behavior of high-frequency features and low-frequency features. 

\begin{figure}[t]
    \centering
    \includegraphics[width=0.5\textwidth]{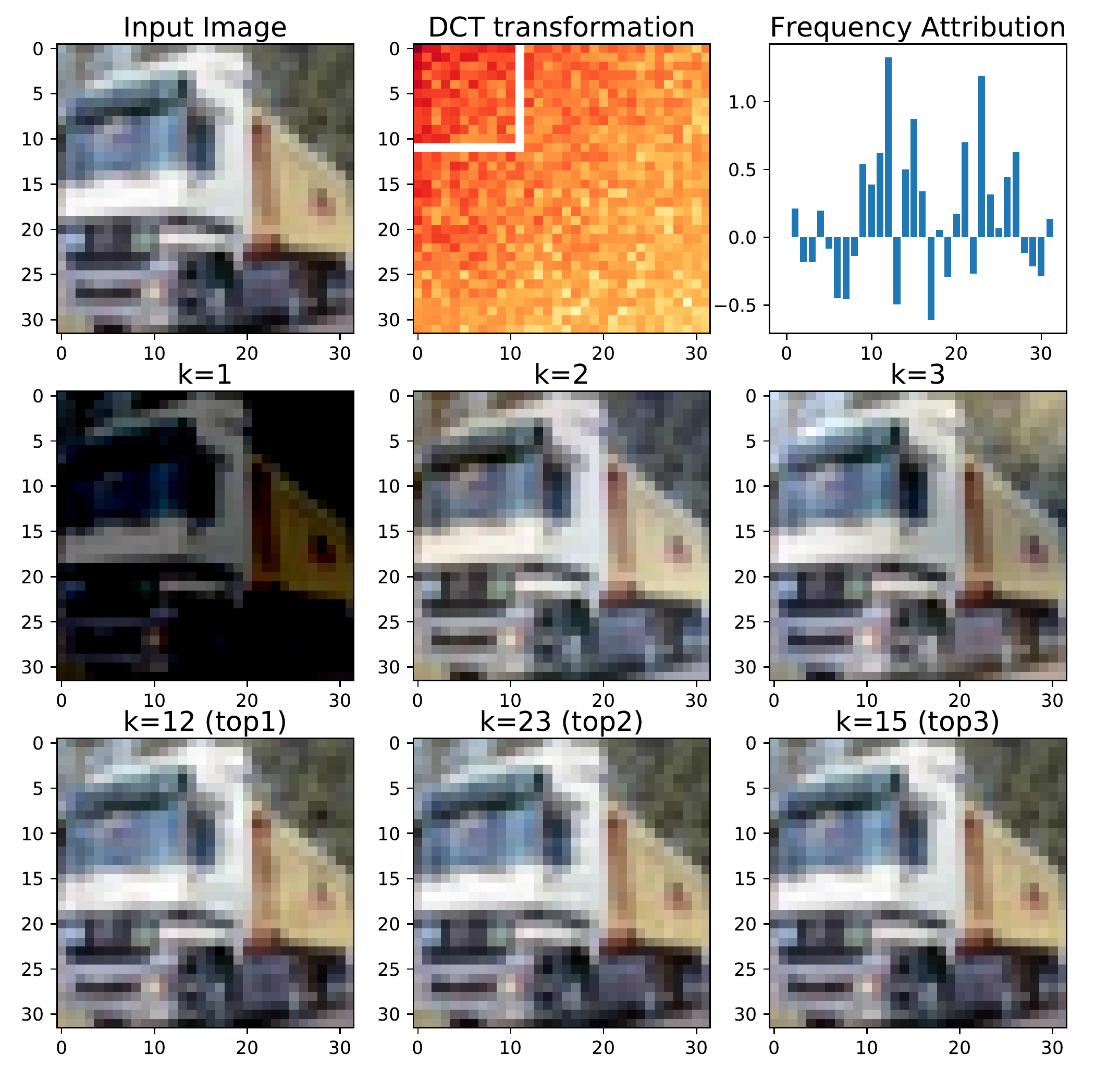}
    \caption{(First Row) Left: the input image. Middle: an example of computing the attribution score for a specific frequency component by ablating the frequency of interest with zeros. Right: the attribution scores for all frequency components in the input on the left. Higher scores denotes higher contribution to the prediction. (Second and Third Row) The input image with the $k-th$ lowest frequency components are ablated. The values of $k$ at the third column are the components with the top1, top2 and top3 attribution scores.}
    \label{rig: cifar_nat}
\end{figure}

With the Occluded Frequency methods, we show an example of computing the frequency attribution on an input image in Cifar-10 which is correctly predicted in Fig \ref{rig: cifar_nat}. We ablate each frequency components from the lowest to the highest to create the bars of contributions on the Frequency Attribution subplot. From the attribution scores, we know that even we assume that the low-frequency component has the strongest correlation with the label since high-frequency components are in-perceptional to humans, it actually takes the less use of the low-frequency components due to low attribution scores in the low-frequency range. We visualize the modified image by OF at a particular frequency. The second row shows image with the lowest, the second lowest and the third lowest frequency components ablated and the the third row shows the modified images when the frequency components with the highest attribution scores are ablated (top1, top2 and top3 respectively). It is very hard for human to capture the difference when the frequency components with highest attribution scores are ablated. A consequence of using features corresponding to the middle and high-frequency range is that by manipulating the middle and high-frequency features, we should expect the change happens to the model's prediction while the human is still not aware of the change made on the input. Images like this are usually considered as adversarial examples. 

\subsection*{Experiment IV: Attribution Shift}
We further extent Experiment III to the entire dataset and we compare the attribution scores of different frequency components on a naturally trained ResNet model and adversarial trained models. There are three robust models trained with the maximum allowed perturbation of $0.25$ and $0.5$ in the $\ell_2$ space and $8/255$ in the $\ell_\infty$ space. The result is shown in Fig \ref{rig: freq_attr}. We plot the average attribution scores on each frequency component for all images in the same classes since we assume images within the same classes should have fewer nuances and more commonalities compared to images of different classes. Despite the lowest frequency component being the most important overall models, Fig \ref{rig: freq_attr} shows that attribution scores on the middle and high-frequency components tend to be zeros or very low and the main share of the attribution scores are in the low-frequency range. It can be seen as the robust model shifts the main share of attribution scores towards the low-frequency range, leaving middle and high-frequency components as insignificant as possible. 

\noindent \textbf{Summary} Finally, we summarize the findings from all experiments. Middle and high-frequency components are the target features for adversarial attackers since models may or may not rely only on the low-frequency features to make the prediction. Adversarial training can be viewed as a frequency selector that the association between the input and its prediction is more tight with low-frequency components, leaving the high-frequency components less relevant for the prediction; therefore, the attacker who aims on the high-frequency features will fail.

\section{Conclusion}
\label{sec: conclusion}
In this paper, We first show a lower bound of the perturbation applied to the image is related to the lowest frequency components, the most perceptional frequency components to human. We then verify that white-box and black-box attackers mainly focus on the high-frequency components in the clean images. We further propose Occluded Frequency as an attribution method to quantify the contribution of different frequency components in the input. Based on the empirical results on the entire dataset, we provide a frequency-based explanation to answer why standard models are not robust: low-frequency features are more robust than the high-frequency features in the input space; therefore robust models are developed to be strongly related to the low-frequency features while standard models rely more on the high-frequency components to make the prediction. 
\section*{Acknowledgement}
We thank Mr. Haohan Wang for providing valuable feedback during the discussion.  

\bibliography{references}

\begin{thebibliography}{28}
\providecommand{\natexlab}[1]{#1}
\providecommand{\url}[1]{\texttt{#1}}
\expandafter\ifx\csname urlstyle\endcsname\relax
  \providecommand{\doi}[1]{doi: #1}\else
  \providecommand{\doi}{doi: \begingroup \urlstyle{rm}\Url}\fi

\bibitem[Bracewell \& Newbold(1986)Bracewell and Newbold]{bracewell1986fourier}
Bracewell and Newbold, R.
\newblock \emph{The Fourier transform and its applications}, volume 31999.
\newblock McGraw-Hill New York, 1986.

\bibitem[{Carlini} \& {Wagner}(2017){Carlini} and {Wagner}]{7958570}
{Carlini}, N. and {Wagner}, D.
\newblock Towards evaluating the robustness of neural networks.
\newblock In \emph{2017 IEEE Symposium on Security and Privacy (SP)}, pp.\
  39--57, 2017.

\bibitem[Cohen et~al.(2019)Cohen, Rosenfeld, and Kolter]{cohen2019certified}
Cohen, J.~M., Rosenfeld, E., and Kolter, J.~Z.
\newblock Certified adversarial robustness via randomized smoothing, 2019.

\bibitem[Engstrom et~al.(2019)Engstrom, Ilyas, Santurkar, and
  Tsipras]{robustness}
Engstrom, L., Ilyas, A., Santurkar, S., and Tsipras, D.
\newblock Robustness (python library), 2019.
\newblock URL \url{https://github.com/MadryLab/robustness}.

\bibitem[Ghorbani et~al.(2017)Ghorbani, Abid, and
  Zou]{ghorbani2017interpretation}
Ghorbani, A., Abid, A., and Zou, J.
\newblock Interpretation of neural networks is fragile, 2017.

\bibitem[Goodfellow et~al.(2014)Goodfellow, Shlens, and
  Szegedy]{goodfellow2014explaining}
Goodfellow, I.~J., Shlens, J., and Szegedy, C.
\newblock Explaining and harnessing adversarial examples, 2014.

\bibitem[Guo et~al.(2018)Guo, Frank, and Weinberger]{guo2018low}
Guo, C., Frank, J.~S., and Weinberger, K.~Q.
\newblock Low frequency adversarial perturbation, 2018.

\bibitem[Guo et~al.(2019)Guo, Gardner, You, Wilson, and Weinberger]{simba}
Guo, C., Gardner, J.~R., You, Y., Wilson, A.~G., and Weinberger, K.~Q.
\newblock Simple black-box adversarial attacks.
\newblock \emph{CoRR}, abs/1905.07121, 2019.
\newblock URL \url{http://arxiv.org/abs/1905.07121}.

\bibitem[He et~al.(2015)He, Zhang, Ren, and Sun]{he2015deep}
He, K., Zhang, X., Ren, S., and Sun, J.
\newblock Deep residual learning for image recognition, 2015.

\bibitem[Ilyas et~al.(2019)Ilyas, Santurkar, Tsipras, Engstrom, Tran, and
  Madry]{ilyas2019adversarial}
Ilyas, A., Santurkar, S., Tsipras, D., Engstrom, L., Tran, B., and Madry, A.
\newblock Adversarial examples are not bugs, they are features, 2019.

\bibitem[Karpathy et~al.(2015)Karpathy, Johnson, and
  Fei-Fei]{karpathy2015visualizing}
Karpathy, A., Johnson, J., and Fei-Fei, L.
\newblock Visualizing and understanding recurrent networks.
\newblock \emph{arXiv preprint arXiv:1506.02078}, 2015.

\bibitem[Koh \& Liang(2017)Koh and Liang]{koh2017understanding}
Koh, P.~W. and Liang, P.
\newblock Understanding black-box predictions via influence functions, 2017.

\bibitem[Kurakin et~al.(2016)Kurakin, Goodfellow, and
  Bengio]{kurakin2016adversarial}
Kurakin, A., Goodfellow, I., and Bengio, S.
\newblock Adversarial examples in the physical world, 2016.

\bibitem[Leino et~al.(2018)Leino, Sen, Datta, Fredrikson, and
  Li]{leino2018influencedirected}
Leino, K., Sen, S., Datta, A., Fredrikson, M., and Li, L.
\newblock Influence-directed explanations for deep convolutional networks,
  2018.

\bibitem[Madry et~al.(2017)Madry, Makelov, Schmidt, Tsipras, and
  Vladu]{aleks2017deep}
Madry, A., Makelov, A., Schmidt, L., Tsipras, D., and Vladu, A.
\newblock Towards deep learning models resistant to adversarial attacks, 2017.

\bibitem[{Narasimha} \& {Peterson}(1978){Narasimha} and {Peterson}]{1094144}
{Narasimha}, M. and {Peterson}, A.
\newblock On the computation of the discrete cosine transform.
\newblock \emph{IEEE Transactions on Communications}, 26\penalty0 (6):\penalty0
  934--936, 1978.

\bibitem[{Papernot} et~al.(2016){Papernot}, {McDaniel}, {Jha}, {Fredrikson},
  {Celik}, and {Swami}]{7467366}
{Papernot}, N., {McDaniel}, P., {Jha}, S., {Fredrikson}, M., {Celik}, Z.~B.,
  and {Swami}, A.
\newblock The limitations of deep learning in adversarial settings.
\newblock In \emph{2016 IEEE European Symposium on Security and Privacy (EuroS
  P)}, pp.\  372--387, 2016.

\bibitem[Rahaman et~al.(2018)Rahaman, Baratin, Arpit, Draxler, Lin, Hamprecht,
  Bengio, and Courville]{rahaman2018spectral}
Rahaman, N., Baratin, A., Arpit, D., Draxler, F., Lin, M., Hamprecht, F.~A.,
  Bengio, Y., and Courville, A.
\newblock On the spectral bias of neural networks, 2018.

\bibitem[Selvaraju et~al.(2019)Selvaraju, Cogswell, Das, Vedantam, Parikh, and
  Batra]{Selvaraju_2019}
Selvaraju, R.~R., Cogswell, M., Das, A., Vedantam, R., Parikh, D., and Batra,
  D.
\newblock Grad-cam: Visual explanations from deep networks via gradient-based
  localization.
\newblock \emph{International Journal of Computer Vision}, 128\penalty0
  (2):\penalty0 336–359, Oct 2019.
\newblock ISSN 1573-1405.
\newblock \doi{10.1007/s11263-019-01228-7}.
\newblock URL \url{http://dx.doi.org/10.1007/s11263-019-01228-7}.

\bibitem[Shafahi et~al.(2019)Shafahi, Najibi, Ghiasi, Xu, Dickerson, Studer,
  Davis, Taylor, and Goldstein]{shafahi2019adversarial}
Shafahi, A., Najibi, M., Ghiasi, A., Xu, Z., Dickerson, J., Studer, C., Davis,
  L.~S., Taylor, G., and Goldstein, T.
\newblock Adversarial training for free!, 2019.

\bibitem[Sharma et~al.(2019)Sharma, Ding, and Brubaker]{Sharma-2019}
Sharma, Y., Ding, G.~W., and Brubaker, M.~A.
\newblock On the effectiveness of low frequency perturbations.
\newblock \emph{CoRR}, abs/1903.00073, 2019.
\newblock URL \url{http://arxiv.org/abs/1903.00073}.

\bibitem[Shrikumar et~al.(2017)Shrikumar, Greenside, and
  Kundaje]{shrikumar2017learning}
Shrikumar, A., Greenside, P., and Kundaje, A.
\newblock Learning important features through propagating activation
  differences, 2017.

\bibitem[Simonyan et~al.(2013)Simonyan, Vedaldi, and
  Zisserman]{simonyan2013deep}
Simonyan, K., Vedaldi, A., and Zisserman, A.
\newblock Deep inside convolutional networks: Visualising image classification
  models and saliency maps, 2013.

\bibitem[Sundararajan et~al.(2017)Sundararajan, Taly, and
  Yan]{sundararajan2017axiomatic}
Sundararajan, M., Taly, A., and Yan, Q.
\newblock Axiomatic attribution for deep networks.
\newblock In \emph{Proceedings of the 34th International Conference on Machine
  Learning-Volume 70}, pp.\  3319--3328. JMLR. org, 2017.

\bibitem[Szegedy et~al.(2013)Szegedy, Zaremba, Sutskever, Bruna, Erhan,
  Goodfellow, and Fergus]{szegedy2013intriguing}
Szegedy, C., Zaremba, W., Sutskever, I., Bruna, J., Erhan, D., Goodfellow, I.,
  and Fergus, R.
\newblock Intriguing properties of neural networks, 2013.

\bibitem[Wang et~al.(2019)Wang, Wu, Yin, and Xing]{Haohan-2020}
Wang, H., Wu, X., Yin, P., and Xing, E.~P.
\newblock High frequency component helps explain the generalization of
  convolutional neural networks.
\newblock \emph{CoRR}, abs/1905.13545, 2019.
\newblock URL \url{http://arxiv.org/abs/1905.13545}.

\bibitem[Yeh et~al.(2018)Yeh, Kim, Yen, and Ravikumar]{yeh2018representer}
Yeh, C.-K., Kim, J.~S., Yen, I. E.~H., and Ravikumar, P.
\newblock Representer point selection for explaining deep neural networks,
  2018.

\bibitem[Zeiler \& Fergus(2013)Zeiler and Fergus]{zeiler2013visualizing}
Zeiler, M.~D. and Fergus, R.
\newblock Visualizing and understanding convolutional networks, 2013.

\end{thebibliography}
\bibliographystyle{icml2019}

\end{document}